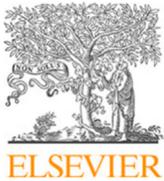
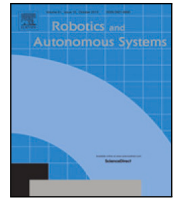

# Cooperative and Asynchronous Transformer-Based Mission Planning for heterogeneous teams of mobile robots

Milad Farjadnasab *, Shahin Sirouspour

*Department of Electrical and Computer Engineering, McMaster University, 1280 Main St W, Hamilton, L8S 4L8, ON, Canada*

## ARTICLE INFO

## ABSTRACT

Cooperative mission planning for heterogeneous teams of mobile robots presents a unique set of challenges, particularly when operating under communication constraints and limited computational resources. To address these challenges, we propose the Cooperative and Asynchronous Transformer-based Mission Planning (CATMiP) framework, which leverages multi-agent reinforcement learning (MARL) to coordinate distributed decision making among agents with diverse sensing, motion, and actuation capabilities, operating under sporadic ad hoc communication. A Class-based Macro-Action Decentralized Partially Observable Markov Decision Process (CMacDec-POMDP) is also formulated to effectively model asynchronous decision-making for heterogeneous teams of agents. The framework utilizes an asynchronous centralized training and distributed execution scheme, enabled by the proposed Asynchronous Multi-Agent Transformer (AMAT) architecture. This design allows a single trained model to generalize to larger environments and accommodate varying team sizes and compositions. We evaluate CATMiP in a 2D grid-world simulation environment and compare its performance against planning-based exploration methods. Results demonstrate CATMiP's superior efficiency, scalability, and robustness to communication dropouts and input noise, highlighting its potential for real-world heterogeneous mobile robot systems. The code is available at https://github.com/mylad13/CATMiP.

## 1. Introduction

Multi-robot systems (MRS) are becoming increasingly prevalent in applications such as search and rescue operations [1], environmental monitoring [2], building and infrastructure inspection [3], and industrial plant management [4]. The coordinated efforts of robots in these systems improve efficiency and adaptability, especially in complex tasks. Particularly, heterogeneous MRS composed of robots with complementary capabilities outperform homogeneous teams in missions requiring diverse sensing and actuation capabilities [5–7].

Coordination in MRS can be centralized or decentralized. Centralized approaches rely on a leader robot or server to issue commands, which can result in high computational and communication loads, vulnerability to single-point failures, and challenges in ensuring consistent communication in real-world scenarios. In contrast, decentralized approaches allow robots to individually make autonomous decisions while implicitly considering the actions of others and changes in the environment. Combining implicit coordination with explicit communication through ad hoc wireless mesh networks enables distributed control strategies that are both scalable and efficient [8,9].

Deep Multi-Agent Reinforcement Learning (MARL) has emerged as a powerful tool for coordinating MRS in dynamic and uncertain environments [10]. Deep MARL enables robots to learn coordination strategies autonomously, bypassing the need for predefined algorithms and heuristics. However, conventional MARL often assumes synchronous decision-making, where agents take new actions at the same time—a condition that is inefficient and impractical for many real-world scenarios. Asynchronous MARL [11] addresses this limitation by enabling agents to make decisions over temporally extended actions, otherwise known as macro-actions [12].

This paper addresses the distributed coordination of heterogeneous mobile robots navigating unknown environments by proposing the Cooperative and Asynchronous Transformer-based Mission Planning (CATMiP) framework. CATMiP is formulated based on the *Class-based Macro-Action Decentralized Partially Observable Markov Decision Process* (CMacDec-POMDP) model, a novel extension of the MacDec-POMDP model [12] for decentralized multi-agent planning that considers varying properties across different agent classes. Our case study involves two robot types-*explorers* and *rescuers*, where the objective is for a rescuer type robot to reach a target with an initially unknown location as fast as possible.

The robots perform collaborative simultaneous localization and mapping (C-SLAM), merging local occupancy grid maps into a shared global map through intermittent communication [13]. Navigation decisions are made in a distributed manner and following a hierarchical two-level control approach. At the high-level, each robot selects a *macro-action*, which is a goal point location on the map within a fixed distance from the robot. This macro-action is sampled from a policy generated by the proposed Asynchronous Multi-Agent Transformer (AMAT) network. The inputs to AMAT are the agents' *macro-observations*, which are class-specific multi-channeled global and local maps in this scenario. At the low-level, path planning and motion control modules generate the robot's immediate action to navigate toward the selected goal.

Section 2 reviews related works and discusses how the identified research gaps are addressed in this paper. Section 3 formally states the problem, introduces the CMacDec-POMDP formulation, and motivates the use of the proposed sequential decision-making approach. Section 4 details the different components of the CATMiP framework, as well as the asynchronous and distributed operation of the robots. The training process and structure of the AMAT policy network are provided in Section 5. Section 6 describes the simulation setup, including macro-observations, macro-actions, and reward structures. Simulation results are presented and analyzed in Section 7. Finally, Section 8 concludes the paper and outlines future directions.

## 2. Related works

Deep reinforcement learning (DRL) has been increasingly used in mobile robotics for exploration and navigation, handling complex tasks in single-agent and multi-agent settings [10,14,15]. Such control strategies are typically divided into end-to-end and two-stage approaches. The end-to-end methods derive control actions directly from sensor data, whereas the two-stage approaches first select target locations using DRL and then employ a separate method for control actions, improving sample efficiency and generalization. Notable recent works have combined high-level DRL-based goal selection with classical path-planning algorithms in single robot scenarios [16–18].

Cooperative multi-robot mission planning has been studied using various deep MARL approaches. Notable works addressing asynchronous multi-robot exploration with homogeneous robots and macro-actions include [19,20]. Tan et al. [19] tackle the challenge of communication dropouts in multi-robot exploration by modeling the problem as a MacDec-POMDP and proposing a DRL solution based on the centralized training and decentralized execution (CTDE) paradigm [21]. CTDE strikes a balance between coordination and scalability by enabling agents to learn from shared experiences during training while acting independently based on local observations during execution. Yu et al. [20] extend the multi-agent proximal policy optimization (MAPPO) algorithm [22] to enable asynchronous CTDE. Their approach enhances coordination efficiency through an attention-based relation encoder, which aggregates feature maps from different agents to capture intra-agent interactions. While these methods demonstrate the effectiveness of macro-actions in asynchronous decision-making, all agents follow the same policy and the unique challenges of planning for heterogeneous multi-robot systems are not considered.

To address heterogeneity, Zhang et al. [23] propose an architecture for asynchronous multi-robot decision-making that combines value function decomposition [24], the MacDec-POMDP framework, and the CTDE paradigm. Their approach utilizes features extracted from both global states and local observations during training. However, during execution, each agent generates macro-actions based solely on feature maps derived from its local observations. This design enables diverse behaviors among agents but restricts the trained model to be used by a fixed team size and composition. Moreover, none of the aforementioned methods allow the trained models to generalize to larger environments. This limitation highlights the need for approaches that prioritize scalability and adaptability in multi-robot mission planning.

To enable agent heterogeneity while maintaining the benefits of parameter sharing, agent indication was formalized in [25]. This method appends an agent-specific indicator signal to the observations, allowing a shared policy network to generate agent-specific actions. Terry et al. [25] demonstrated that parameter sharing can be effectively applied to heterogeneous observation and action spaces while still achieving optimal policies. This idea is used in the Multi-Agent Transformer (MAT) architecture [26] as well, where positional encoding that appears in the original transformer [27] are replaced by agent indication.

Wen et al. [26] introduced MAT alongside a novel MARL training paradigm that achieves linear time complexity and guarantees monotonic performance improvement by leveraging the *multi-agent advantage decomposition theorem* [28]. This theorem suggests that joint positive advantage can be achieved by sequentially selecting local actions rather than searching the entire joint action space simultaneously. Thus, cooperative MARL can be reformulated as a sequence modeling problem, where the objective is to map a sequence of agent observations to a sequence of optimal agent actions.

In MAT, the attention mechanism [27] in the encoder captures the inter-agent relationships within the sequence of observations, and the decoder autoregressively generates actions by considering the input sequence's latent representation. The transformer model's ability to process flexible sequence lengths enables generalization to different team sizes without treating varying agent numbers as separate tasks. This property allows a single trained model to scale to teams with more or less agents than those encountered during training.

Building on MAT, our work introduces the Asynchronous Multi-Agent Transformer (AMAT) network. We develop a new asynchronous centralized training and asynchronous distributed execution scheme tailored for heterogeneous teams. Specialized agent class policies are learned through *agent class encodings*, which differentiate macro-observations across agent classes and enable the network to generate corresponding macro-actions. This design enhances the model's generalizability to larger teams with varying compositions of heterogeneous agents. Additionally, we employ an adaptive pooling layer during global feature extraction from macro-observations, allowing the model to efficiently scale to larger environment sizes without compromising performance.

## 3. Preliminaries and problem formulation

### 3.1. Problem statement

This paper addresses the design of distributed controllers for a heterogeneous team of mobile robots performing a cooperative mission in an unknown environment. Specifically, we focus on an indoor search and target acquisition scenario involving two agent classes: explorers and rescuers. The mission objective is for a rescuer agent to reach a target with an initially unknown location as quickly as possible. To achieve this, control policies must exploit the diverse capabilities of the team, encouraging specialized behaviors for each agent class. For instance, explorer robots, being faster and more agile, are tasked with rapidly mapping the environment and locating the target. In contrast, rescuer robots, though slower, have the capability to engage with the target once its location is known.

To solve this problem, we propose a hierarchical control approach that combines high-level decision-making and low-level motion control for effective navigation. First, a goal location within a localized area centered on the robot is selected by the high-level decision-making module; then, the local planner generates motion commands to ensure smooth movement and obstacle avoidance en route to the goal.

*3.2. The CMacDec-POMDP model*

We formalize this approach as a Class-based Macro Action Decentralized Partially Observable Markov Decision Process (CMacDec-POMDP), a novel extension of the MacDec-POMDP framework [12] that incorporates varying properties across different agent classes. Assuming a team of $N$ agents with $C \leq N$ different classes, the problem is formalized as the tuple $\langle \mathcal{I}, C, S, \{M^c\}, \{A^c\}, P, \{R^c\}, \{\zeta^c\}, \{Z^c\}, \{\Omega^c\}, \{O^c\}, \gamma, h \rangle$, where

- $\mathcal{I} = \{1, \ldots, N\}$ is a finite set of agents;
- $C = \{1, \ldots, C\}$ is a finite set of agent classes, with $C(i) \in C$ indicating the class of agent $i \in \mathcal{I}$;
- $S$ is the global state space;
- $M^c$ is a finite set of macro-actions (MAs) for agents of class $c$. The set of joint MAs is then $M = \times_i M^{C(i)}$;
- $A^c$ is a finite set of (primitive) actions for agents of class $c$. The set of joint actions is then $A = \times_i A^{C(i)}$;
- $P : S \times A \times S \to [0, 1]$ is a state transition probability function, indicating the probability of transitioning from state $s \in S$ to state $s' \in S$ when the agents are taking the joint action $\bar{a} \in A$. In other words, $P(s, \bar{a}, s') = Pr(s'|\bar{a}, s)$;
- $R^c : S \times A \to \mathbb{R}$ is the agent-class-specific reward function, with $R^{C(i)}(s, \bar{a})$ being the reward an agent $i$ of class $c$ receives when the joint action $\bar{a}$ is executed in state $s$;
- $\zeta^c$ is a finite set of macro-observations (MOs) for agents of class $c$. The set of joint MOs is then $\zeta = \times_i \zeta^{C(i)}$;
- $Z^c : \zeta^c \times M^c \times S \to [0, 1]$ is the MO probability function for agents of class $c$, indicating the probability of the agent receiving the MO $z^i \in \zeta^{C(i)}$ given MA $m^i \in M^{C(i)}$ is in progress or has completed and the current state is $s' \in S$. In other words, $Z^{C(i)}(z^i, m^i, s') = Pr(z^i|m^i, s')$;
- $\Omega^c$ is a finite set of observations for agents of class $c$. The set of joint observations is then $\Omega = \times_i \Omega^{C(i)}$;
- $O^c : \Omega^c \times A^c \times S \to [0, 1]$ is the observation probability function for agents of class $c$, indicating the probability of agent $i$ receiving the observation $o^i \in \Omega^{C(i)}$ when the current state is $s' \in S$ after the agents have taken the joint action $\bar{a} \in A$. In other words, $O^{C(i)}(o^i, \bar{a}, s') = Pr(o^i|\bar{a}, s')$;
- $\gamma \in [0, 1]$ is the discount factor;
- and $h$ is the horizon, the number of steps in each episode.

We denote primitive time steps as $t = 0, 1, 2, \ldots, h$, and global macro-steps as $\tau = 0, 1, 2, \ldots, T$, where the final macro-step index $T$ may vary across episodes. At each primitive time step $t$, an agent $i \in \mathcal{I}$ either belongs to the set of *active* agents $\mathcal{A}_t \subseteq \mathcal{I}$, and therefore samples a new macro-action from its high-level policy $\mu^i : H_M^i \times M^{C(i)} \to [0, 1]$, or to the set of *busy* agents $\mathcal{B}_t = \mathcal{I} \setminus \mathcal{A}_t$, still executing a previously assigned macro-action. The global macro-step index $\tau$ is incremented only when at least one agent is active at time $t$, i.e., when $\mathcal{A}_t \neq \emptyset$. We define a mapping from global macro-step indices to primitive time steps as:

$$\psi : \{0, 1, \ldots, T\} \to \{0, 1, \ldots, h\}, \quad \psi(\tau) = t_\tau,$$

where $t_\tau$ is the primitive time step at which macro-step $\tau$ occurs. Additionally, each agent $i \in \mathcal{I}$ maintains its own macro-step counter $\tau^i = 0, 1, 2, \ldots, T^i$, which is incremented only when the agent becomes active and selects a new macro-action. We define the corresponding agent-specific mapping:

$$\psi^i : \{0, 1, \ldots, T^i\} \to \{0, 1, \ldots, h\}, \quad \psi^i(\tau^i) = t_{\tau^i},$$

where $t_{\tau^i}$ denotes the primitive time step at which agent $i$ initiated its $\tau^i$-th macro-action. To relate each agent's local macro-step to the global macro-step sequence, we define the mapping:

$$\kappa^i : \{0, 1, \ldots, T^i\} \to \{0, 1, \ldots, T\}, \quad \kappa^i(\tau^i) = \tau,$$

**Table 1**
Temporal mappings between primitive time, global macro-steps, and per-agent macro-steps.

| Mapping | Domain → Codomain | Description |
| --- | --- | --- |
| $\psi(\tau)$ | $\{0, \ldots, T\} \to \{0, \ldots, h\}$ | Global macro-step $\tau \mapsto$ primitive time $t_\tau$ |
| $\psi^i(\tau^i)$ | $\{0, \ldots, T^i\} \to \{0, \ldots, h\}$ | Agent $i$'s macro-step $\tau^i \mapsto$ primitive time $t_{\tau^i}$ |
| $\kappa^i(\tau^i)$ | $\{0, \ldots, T^i\} \to \{0, \ldots, T\}$ | Agent $i$'s macro-step $\tau^i \mapsto$ global macro-step $\tau$ |

which maps the agent's $\tau^i$-th macro-step to the corresponding global macro-step index $\tau$ at which it occurred. These mappings are summarized in Table 1.

A macro-action is a temporally extended action (also referred to as an *option* or *skill*) represented as $m^i = \langle \beta_{m^i}, \mathcal{I}_{m^i}, \pi_{m^i} \rangle$, where $\beta_{m^i} : H_A^i \to [0, 1]$ is a stochastic termination condition based on the primitive action-observation history $H_A^i = (a_0^i, o_0^i, \ldots, a_t^i, o_t^i)$, $\mathcal{I}_{m^i} \subset H_M^i$ is the initiation condition that determines whether the MA can be started based on the macro-observation-action history $H_M^i = (z_0^i, m_0^i, \ldots, z_{\tau^i}^i, m_{\tau^i}^i)$, and $\pi_{m^i} : H_A^i \times A^{C(i)} \to [0, 1]$ is the low-level control policy that generates primitive actions $a^i$ to execute the macro-action.

The underlying Dec-POMDP is used to generate primitive transitions and rewards, but the low-level policy $\pi_{m^i}$ of agent $i$ is determined by the MA obtained via the high-level policy $\mu^i$. Access to the full model of the underlying Dec-POMDP is not necessary, as the MAs are assumed to be simulated in an environment close enough to the real-world domain. This allows all evaluations to be conducted in the simulator through sampling [12].

The CMaCDec-POMDP model is flexible enough to model heterogeneous multi-agent teams operating under either cooperative or mixed cooperative-competitive objectives, depending on how the reward functions are specified. In general, each agent seeks to maximize its own *expected total return*:

$$v^i(\bar{\mu}) := \mathbb{E}_{\bar{\mu}, \bar{\pi}} \left[ \sum_{t=0}^{h-1} \gamma^t r_t^i \,\middle|\, s_0 \right], \tag{1}$$

where $r_t^i = R^{C(i)}(s_t, \bar{a}_t)$ may differ between agent classes. In our setting, we focus on fully cooperative teams and design team-oriented reward functions, defining the overall *team objective* as the sum of individual returns. The goal is to find a joint high-level policy $\bar{\mu} = (\mu^1, \ldots, \mu^N)$ that maximizes this objective:

$$\bar{\mu}^* = \underset{\bar{\mu}}{\operatorname{argmax}} \quad \mathbb{E}_{\bar{\mu}, \bar{\pi}} \left[ \sum_{i=1}^{N} v^i(\bar{\mu}) \right]. \tag{2}$$

*3.3. Sequential approach to the solution*

In the general form of the CMacDec-POMDP model, the agents do not observe the global state and their high-level policy is a function of local macro-observation-action histories. Following the centralized training with decentralized execution (CTDE) paradigm, we define the *joint macro-observation value function* at every macro-step $\tau$ as a function of joint MOs $\bar{z}_\tau = (z_\tau^1, \ldots, z_\tau^N)$ as:

$$V_{\bar{\mu}}(\bar{z}_\tau) := \mathbb{E}_{\bar{\mu}, \bar{\pi}} \left[ \sum_{t=t_\tau}^{h-1} \sum_{i=1}^{N} \gamma^{t-t_\tau} r_t^i \,\middle|\, \bar{z}_\tau \right], \tag{3}$$

which is the expected sum of individual returns starting from macro-step $\tau$ and following the joint policy $\bar{\mu}$. The local macro-observation value function simply represents the expected total return from a single agent, defined as:

$$V_{\bar{\mu}}^i(\bar{z}_\tau) := \mathbb{E}_{\bar{\mu}, \bar{\pi}} \left[ \sum_{t=t_\tau}^{h-1} \gamma^{t-t_\tau} r_t^i \,\middle|\, \bar{z}_\tau \right]. \tag{4}$$

At each global macro-step $\tau$, only the active subset of agents $\mathcal{A}_{t_\tau}$ select new macro-actions. We define the *multi-agent macro-observation-action value function* for an arbitrary ordered subset of active agents

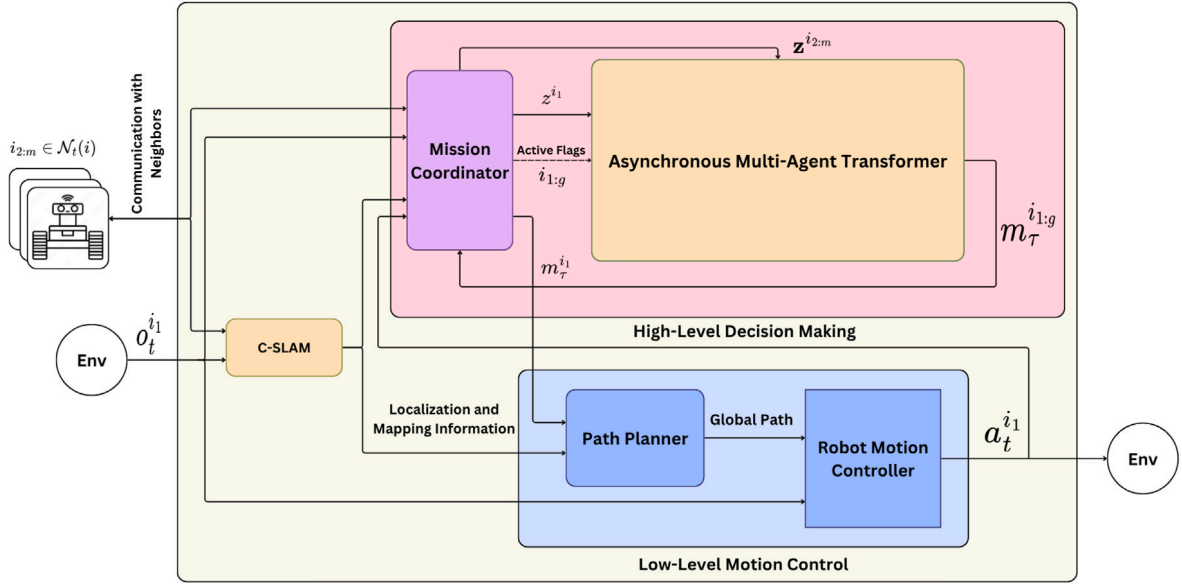

**Fig. 1.** Workflow of the CATMiP framework during the execution phase. The robots communicate to share mission information and their embedded macro-observations to make high-level navigation decisions in a distributed manner, while a local motion controller generates the immediate action for each robot.

$i_{1:k} = \{i_1, \ldots, i_k\} \subset \mathcal{A}_{t_\tau}$ as:

$$Q_{\bar{\mu}}^{i_{1:k}}(\bar{z}_\tau, m_\tau^{i_{1:k}}) := \mathbb{E}_{\bar{\mu}, \bar{\pi}} \left[ \sum_{t=t_\tau}^{h-1} \sum_{i=1}^{N} \gamma^{t-t_\tau} r_t^i \,\middle|\, \bar{z}_\tau, m_\tau^{i_{1:k}} \right], \quad (5)$$

which represents the expected sum of returns when $i_{1:k}$ take the MAs $m_\tau^{i_{1:k}}$. Note that for $k = 0$, this becomes the joint macro-observation value function. Given another arbitrary subset of active agents $j_{1:l} = \{j_1, \ldots, j_l\} \subset \mathcal{A}_{t_\tau}$ disjoint from $i_{1:k}$, we can define the *multi-agent macro-advantage function* as:

$$A_{\bar{\mu}}^{i_{1:k}}(\bar{z}_\tau, m_\tau^{j_{1:l}}, m_\tau^{i_{1:k}}) := Q_{\bar{\mu}}^{j_{1:l}, i_{1:k}}(\bar{z}_\tau, m_\tau^{j_{1:l}}, m_\tau^{i_{1:k}}) - Q_{\bar{\mu}}^{j_{1:l}}(\bar{z}_\tau, m_\tau^{j_{1:l}}), \quad (6)$$

which quantifies the contribution of the subset $i_{1:k}$ of agents to the total return by taking the MAs $m_\tau^{i_{1:k}}$, once $j_{1:l}$ have taken the MAs $m_\tau^{j_{1:l}}$. Again, for $l = 0$, the macro-advantage function assesses this contribution with respect to the baseline joint macro-observation value of the whole team.

The *multi-agent advantage decomposition theorem* [28] provides a principled foundation for optimizing joint multi-agent policies via sequential per-agent updates. Here, we extend it to the macro-level, stating that in every global decision step $\tau$, given an arbitrary fixed ordering of active agents $i_{1:g(\tau)}$, the following equation always holds:

$$A_{\bar{\mu}}^{i_{1:g(\tau)}}(\bar{z}_\tau, m^{i_{1:g(\tau)}}) = \sum_{l=1}^{g(\tau)} A_{\bar{\mu}}^{i_l}(\bar{z}_\tau, m^{i_{1:l-1}}, m^{i_l}). \quad (7)$$

This means that as long as each agent $i_l$ in the ordered subset $i_{1:g(\tau)}$ chooses an MA $m^{i_l}$ with positive advantage conditioned on the joint MOs and the MAs $m^{i_{1:l-1}}$ of its predecessors, the team's overall multi-agent macro-advantage is positive. Consequently, maximizing each agent's local macro-advantage in sequence is equivalent to maximizing the joint team objective, guaranteeing *monotonic performance improvement* during training.

Moreover, this sequential approach reduces the complexity of policy optimization from *exponential* to *linear* in the number of agents: at each macro-step, macro-actions are chosen one at a time, each conditioned only on prior decisions and shared context, rather than searching over the entire joint macro-action space.

Our Asynchronous Multi-Agent Transformer (AMAT) network is explicitly designed to leverage this sequential approach. AMAT employs an autoregressive decoder, using masked self-attention to generate each agent's MA conditioned on both a latent representation of its MO and the MAs of preceding agents in the decision order. This architecture efficiently captures inter-agent dependencies while maintaining the theoretical guarantees of sequential advantage updates, enabling scalable and effective high-level policy learning for distributed cooperation in large, heterogeneous multi-agent teams.

## 4. Framework architecture

The Cooperative and Asynchronous Transformer-based Mission Planning (CATMiP) framework provides a unified solution for scalable, robust cooperation in heterogeneous multi-robot teams operating under communication constraints. The overall architecture of CATMiP, shown in Fig. 1, comprises *three* key modules: the **C-SLAM module**, which provides global mapping and localization; the **High-Level Decision Making module**, which determines macro-actions for agents and handles inter-agent communication; and the **Low-Level Motion Control module**, which translates macro-actions into motion commands for navigation. To enable efficient and scalable operation, CATMiP employs an *asynchronous centralized training* process in a simulation environment to learn agent policies and an *asynchronous distributed execution* scheme to deploy these policies onboard the robots during real-time missions. The following subsections first describe the formation of time-varying communication neighborhoods and the distributed coordination of the robots during a mission, and then detail each module of the framework.

### 4.1. Distributed operation in dynamic communication neighborhoods

Robots share information in a distributed manner during the mission, forming time-varying communication neighborhoods through a dynamic mobile ad hoc network [29]. Each robot in the network functions both as a communication endpoint and as a router, forwarding information on behalf of others when direct communication is not possible. This allows for robust, decentralized information sharing even when robots are not within each other's immediate communication range, provided a path of intermediate nodes exists. At each time step $t$, the communication topology is modeled as a dynamic graph $\mathcal{G}_t = (\mathcal{I}, \mathcal{E}_t)$, where vertices $\mathcal{I}$ represent the set of agents, and edges $\mathcal{E}_t \subseteq \mathcal{I} \times \mathcal{I}$ represent communication links between them. The probability of a communication link $E_{ij} \in \mathcal{E}_t$ existing between agents $i$ and $j$ depends on their distance $d_{ij}$, defined as:

$$Pr(E_{ij}) = e^{-d_{ij}^2/\sigma^2}, \tag{8}$$

where $\sigma$ is a decay parameter controlling how quickly the communication probability decreases with distance. This probabilistic model is adapted from [30]. The set of communication neighborhoods (the connected components of the graph) at time $t$ is defined as

$$\mathbb{N}_t = \left\{ \mathcal{N}_t^{(1)}, \mathcal{N}_t^{(2)}, \ldots, \mathcal{N}_t^{(k_t)} \right\},$$

where each $\mathcal{N}_t^{(k)} \subseteq \mathcal{I}$ is a maximal set of agents who are mutually reachable in $\mathcal{G}_t$, and $\mathcal{N}_t^{(k)} \cap \mathcal{N}_t^{(l)} = \emptyset$ for $k \neq l$. Each agent $i \in \mathcal{I}$ belongs to exactly one communication group at time $t$, so we can define a mapping:

$$\mathcal{N}_t(i) = \text{the unique } \mathcal{N}_t^{(k)} \in \mathbb{N}_t \text{ such that } i \in \mathcal{N}_t^{(k)}.$$

At every macro-step $\tau$, *active* agents within the same communication neighborhood ($i$), denoted as $\mathcal{A}_{t_\tau}^{(i)} := \mathcal{A}_{t_\tau} \cap \mathcal{N}_{t_\tau}^{(i)}$, select a *temporary coordinator*. The selection can be based on a fixed protocol, such as choosing the robot with lowest ID, most direct communication links, or highest computational capacity. The coordinator robot facilitates distributed high-level decision-making by aggregating macro-observation embeddings from all agents in the neighborhood $\mathcal{N}_{t_\tau}^{(i)}$, processing them onboard through its trained policy network, and distributing the generated macro-actions to all active agents in the set $\mathcal{A}_{t_\tau}^{(i)}$. Fig. 1 shows the overview of the process, while the details are provided in Section 5. This mechanism enables scalable coordination without requiring a permanent centralized controller, adapting to the dynamic, time-varying nature of the communication network.

*4.2. Collaborative Simultaneous Localization and Mapping (C-SLAM)*

This module enables robots to collaboratively build a shared global occupancy grid map. To ensure scalability in large teams, we may employ a fully decentralized C-SLAM method such as Swarm-SLAM [13], which is specifically designed for resource efficiency and sparse communication. Instead of requiring all robots to send full maps or trajectory data to a central server, map merging and optimization are performed locally within small communication neighborhoods, and only compact descriptors or key loop closure information are exchanged.

During each rendezvous, an elected robot within the neighborhood performs the necessary computation for map merging and pose graph optimization. This reduces redundant computation and ensures that no single robot or external server becomes a bottleneck. As a result of the decentralized and neighborhood-based approach, robots in different communication neighborhoods may temporarily use different versions of the shared map to make decisions and each robot bases its high-level decision-making on the most recent map available within its current communication group. When neighborhoods merge through robot rendezvous, maps are reconciled and updated using the latest aggregated information, ensuring eventual consistency across the team. Extensive experiments in Swarm-SLAM [13] demonstrate that this decentralized, resource-aware approach maintains mapping accuracy and real-time performance as the number of robots increases, even in challenging environments with intermittent connectivity. During our centralized training process, all robots are assumed to be always connected and having access to the same shared map. The occupancy grid map is then passed onto the high-level decision making and low-level motion control modules.

*4.3. High-level decision making*

The High-Level Decision-Making module determines macro-actions for each agent and enables coordination by leveraging information from robot sensory data, the C-SLAM module, and inter-agent communication.

During the centralized training phase, the **Mission Coordinator** sub-module aggregates mission-related information and forms all agents' MOs, denoted as $\bar{z}$. The Mission Coordinator maintains the action-observation history $\bar{H}_A$ and the macro-action-macro-observation history $\bar{H}_M$ of all agents. It manages *agent activation* and determines new MAs by passing $\bar{z}$ through the **Asynchronous Multi-Agent Transformer (AMAT)** network, detailed in Section 5. The selected MAs are then sent to the Low-Level Motion Control module as navigation goals for execution.

In the distributed execution phase, the Mission Coordinator operates locally on each robot, managing agent activation independently by maintaining local $H_A^i$ and $H_M^i$. It also facilitates the exchange of MOs and MAs between communicating agents. This localized approach ensures efficient real-time decision making while maintaining coordination across the team.

*4.4. Low-level motion control*

The Low-Level Motion Control module generates the robot's primitive action $a_t^i$ at each time step $t$ by integrating a path planner and a motion controller. The path planner determines a collision-free path to the global goal specified by the robot's current MA $m^i$. A path-finding algorithm, such as A* search [31], can be used to compute the shortest path on the occupancy grid map, represented as a sequence of waypoints.

To refine the planned path, the motion controller employs a local planner [32], which optimizes the robot's trajectory based on the selected path and real-time sensory data. The local planner ensures that the resulting trajectory adheres to the robot's motion constraints, avoids dynamic and static obstacles, and minimizes execution time. The optimized trajectory is then translated into low-level motion control commands, producing the primitive action $a_t^i$.

## 5. Asynchronous Multi-Agent Transformer (AMAT)

This section details the structure and different components of the AMAT network and its use during asynchronous centralized training and distributed execution.

The AMAT network, illustrated in Fig. 2, consists of four components: **Macro-Observations Embedder**, **Encoder**, **Macro-Actions Embedder**, and **Decoder**. AMAT transforms a sequence of MOs $\{z^{i_l}\}_{l=1}^{m}$ from a subset $i_{1:m}$ of agents into a sequence of MAs $\{m^{i_l}\}_{l=1}^{g}$ corresponding to the active subset $i_{1:g}$ ($g \leq m$). During centralized training, $i_{1:m}$ represents the complete agent set $\mathcal{I} = \{1, \ldots, N\}$, while in distributed execution, it refers to the agents in the neighborhood $\mathcal{N}^{(i)} = \{i_1, \ldots, i_m\}$. In both cases, the sets are ordered to put the active agents $i_{1:g}$ first.

*5.1. Asynchronous centralized training*

Centralized training is conducted in simulation, across multiple parallel environments and over many episodes. At the start of each episode, every agent $i \in \mathcal{I}$ selects an initial MA $m_0^i$ based on its initial MO $z_0^i$. At every time step $t$ of a training episode, the agents execute the joint (primitive) action $\bar{a}_t$ generated by their low-level motion control module, and each collect a class-specific state–action dependent reward, jointly denoted as $\bar{R}(s_t, \bar{a}_t)$. The primitive rewards received by agent $i$ during the execution of its $\tau^i$th MA are accumulated and stored as $\mathcal{R}_{\tau^i}^i = \sum_{t=t_{\tau^i}}^{t_{\tau^i+1}-1} R^{C(i)}(s_t, \bar{a}_t)$.

At every global macro-step $\tau$, agents are reordered with the permutation $perm_\tau : \mathcal{I} \to (i_1, \ldots, i_{g(\tau)}, i_{g(\tau)+1}, \ldots, i_N) = \{i_l\}_{l=1}^{N}$, starting with a fixed arbitrary ordering of $g(\tau) = |\mathcal{A}_{t_\tau}|$ active agents and followed by a fixed arbitrary ordering of busy agents. A corresponding sequence of MOs $\{z_\tau^{i_l}\}_{l=1}^{N}$ are formed, converted into MO-embeddings $\{\mathbf{z}_\tau^{i_l}\}_{l=1}^{N}$, and fed as the input to the encoder of the AMAT network, resulting in a sequence of latent representations of MOs of active agents, denoted as $\{\hat{\mathbf{z}}_\tau^{i_l}\}_{l=1}^{g(\tau)}$. These latent representations are used by the encoder's output layer to obtain value function estimates $\{V(\hat{\mathbf{z}}_\tau^{i_l})\}_{l=1}^{g(\tau)}$, and by the decoder to calculate new MAs $\{m_\tau^{i_l}\}_{l=1}^{g(\tau)}$ for the active agents. The current sequence of macro-observations of all agents $\{z_\tau^{i_l}\}_{l=1}^{N}$, the permutation $perm_\tau$, per-agent macro-step counters $\{\tau^{i_l}\}_{l=1}^{g(\tau)}$ that correspond to the global macro-step $\tau$, as well as the macro-actions, accumulated rewards, and value function estimates from the previous activation instance of active agents, $\{m_{\tau^{i_l}-1}^{i_l}, \mathcal{R}_{\tau^{i_l}-1}^{i_l}, V(\hat{\mathbf{z}}_{\tau^{i_l}-1}^{i_l})\}_{l=1}^{g(\tau)}$, are stored into a replay buffer.

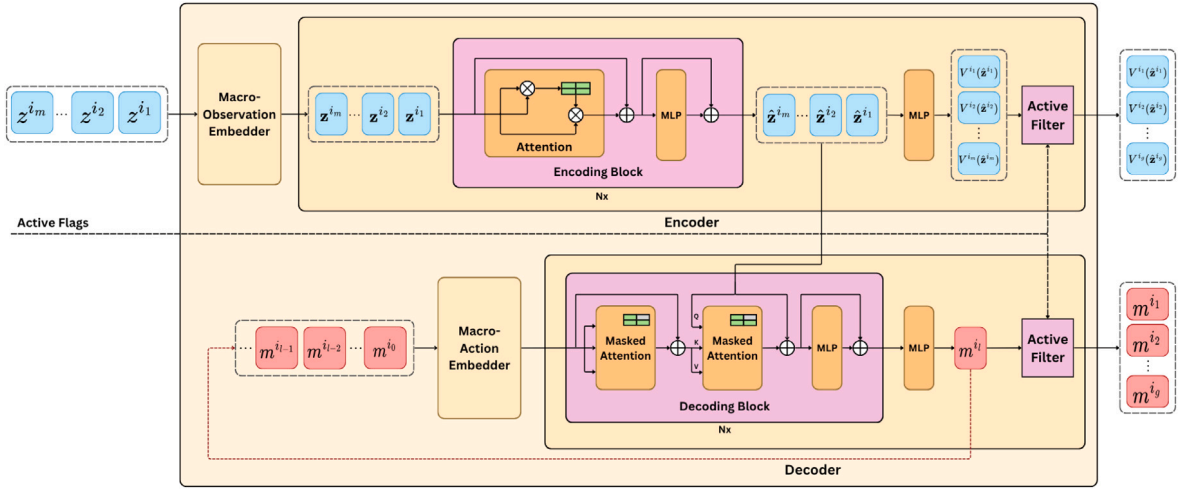

**Fig. 2.** Centralized macro-action inference as a part of the training process of AMAT. During distributed execution, the coordinator robot receives macro-observation embeddings from connected agents and transmits the newly obtained macro-actions back to them.

At the end of each training episode, the stored transitions are aligned by macro-steps to form the experience trajectory $\left(\{z_\tau^{i_l}\}_{l=1}^{N}, \{m_\tau^{i_l}, \mathcal{R}_{\tau^{i_l}}^{i_l}, V(\hat{\mathbf{z}}_\tau^{i_l})\}_{l=1}^{g(\tau)}\right)$, and used to optimize the AMAT network parameters through backpropagation.

It should be noted that during training, *busy* agents delay their transition to the *active* state for a random number of time steps, chosen between a predefined minimum $t_{min}$ and maximum $t_{max}$. If the minimum delay has elapsed for at least two agents, those agents activate together and update their MAs with a single pass through the AMAT network. If the maximum delay is reached for any agent and no other agents have finished their MA and passed their minimum delay, that agent activates alone. This mechanism not only reduces the total number of experience trajectories gathered during training and decreases the computation cost, but also helps simulate the unforeseen variance in the time it takes to execute an MA in real-world conditions, alleviating the sim-to-real gap [20].

After training is completed, the resulting fully trained network is deployed on each robot's local hardware for decentralized or distributed execution, enabling real-time inference of macro-actions during the mission. The pseudocode of the centralized training process is presented in Algorithm 1.

### 5.2. Macro-observations embedder

During centralized training, this module transforms a sequence of MOs $(z^{i_1}, \ldots, z^{i_m})$ into a sequence of embeddings $(\mathbf{z}^{i_1}, \ldots, \mathbf{z}^{i_m})$ to be used as input tokens for the encoder. In the distributed execution phase, each agent $i_l \in \mathcal{N}^{(i)}$ locally processes its MO $z^{i_l}$ using the trained embedder, then transmits the resulting embedding $\mathbf{z}^{i_l} \in \mathbb{R}^d$ to its neighborhood's coordinator. This approach reduces network traffic by transmitting compact embeddings instead of high-dimensional MOs, which include local and global maps.

The architecture, shown in Fig. 3, uses two separate convolutional neural networks (CNNs) to extract features from multi-channeled global and local maps. The CNN processing the global map uses a single

---

**Algorithm 1** Asynchronous Centralized Training of CATMiP

**Input:** Agents $\mathcal{I}$, episodes $K$, steps per episode $h$, minibatch size $b$
**Initialize:** Encoder parameters $\phi_0$, Decoder parameters $\theta_0$, Replay Buffer $B$

1: **for** $k = 0, 1, \ldots, K-1$ **do**
2:     Initialize the environment
3:     Form initial macro-observation $z_0^i$, and obtain initial macro-action $m_0^i$ and value function estimate $V(\hat{\mathbf{z}}_0^i)$ for $i \in \mathcal{I}$
4:     Store $\{z_0^i\}_{i \in \mathcal{I}}$ in $B$
5:     Set $t \leftarrow 0$, $\tau \leftarrow 1$, $\tau^i \leftarrow 1$ for all $i \in \mathcal{I}$
6:     Set all agents to busy at $t = 0$, i.e., $\mathcal{A}_0 = \emptyset$
7:     **while** episode not done **and** $t < h$ **do**
8:         **if** $\mathcal{A}_t \neq \emptyset$ **then**
9:             Reorder agents with the permutation $perm_\tau : \mathcal{I} \rightarrow (i_1, \ldots, i_{g(\tau)}, i_{g(\tau)+1}, \ldots, i_N)$ (active first), with $g(\tau) = |\mathcal{A}_t|$
10:            Form macro-observation sequence $\{z_\tau^{i_l}\}_{l=1}^{N}$
11:            Obtain latent MO embeddings $\{\hat{\mathbf{z}}_\tau^{i_l}\}_{l=1}^{g(\tau)}$ for active agents
12:            Compute value estimates $\{V(\hat{\mathbf{z}}_\tau^{i_l})\}_{l=1}^{g(\tau)}$ for active agents using the value head of AMAT's encoder
13:            Autoregressively generate MAs $\{m_\tau^{i_l}\}_{l=1}^{g(\tau)}$ for active agents using AMAT's decoder
14:            Set $\mathcal{R}_{\tau^{i_l}-1}^{i_l} \leftarrow \mathcal{R}^{i_l}$
15:            Store $\left(\{z_\tau^{i_l}\}_{l=1}^{N}, \{m_{\tau^{i_l}-1}^{i_l}, \mathcal{R}_{\tau^{i_l}-1}^{i_l}, V(\hat{\mathbf{z}}_{\tau^{i_l}-1}^{i_l}), \tau^{i_l}\}_{l=1}^{g(\tau)}, \tau, perm_\tau\right)$ in $B$
16:            Reset $\mathcal{R}^{i_l} \leftarrow 0$ for each $i_l \in \mathcal{A}_t$
17:            Increment counters: $\tau \leftarrow \tau + 1$, $\tau^{i_l} \leftarrow \tau^{i_l} + 1$ for all $i_l \in \mathcal{A}_t$
18:         **end if**
19:         All agents execute joint primitive action $\bar{a}_t$ and receive $R^{C(i)}(s_t, \bar{a}_t)$
20:         **for** each $i \in \mathcal{I}$ **do**
21:            Accumulate macro-reward: $\mathcal{R}^i \leftarrow \mathcal{R}^i + \gamma^{t-t_{\tau^i}} R^{C(i)}(s_t, \bar{a}_t)$
22:         **end for**
23:         $t \leftarrow t + 1$
24:     **end while**
25:     Align the experiences of all agents in the buffer $B$ with global macro-steps, forming $T$ experience trajectories $\left(\{z_\tau^{i_l}\}_{l=1}^{N}, \{m_{\tau^{i_l}}^{i_l}, \mathcal{R}_{\tau^{i_l}}^{i_l}, V(\hat{\mathbf{z}}_{\tau^{i_l}}^{i_l})\}_{l=1}^{g(\tau)}\right)$
26:     Sample minibatch of $b$ experiences from $B$
27:     Update parameters: $\phi_{k+1}, \theta_{k+1} \leftarrow$ minimize $L_{\text{Encoder}}(\phi) + L_{\text{Decoder}}(\theta)$ (Eqs. (9) and (10))
28: **end for**

---

2D convolution layer with 32 output channels and a kernel size of 7, padding of 3, and stride of 1, followed by an adaptive max pooling layer that scales global feature maps of any $S \times S$ dimensions to fixed $G \times G$ zones, enabling the model's operation in varying environment sizes. The CNN processing the local map is composed of two layers with output channels of 64 and 32, respectively, and both with kernel sizes of 3, padding of 1, and stride of 1. The features are then flattened, concatenated, and processed by a multi-layer perceptron (MLP).

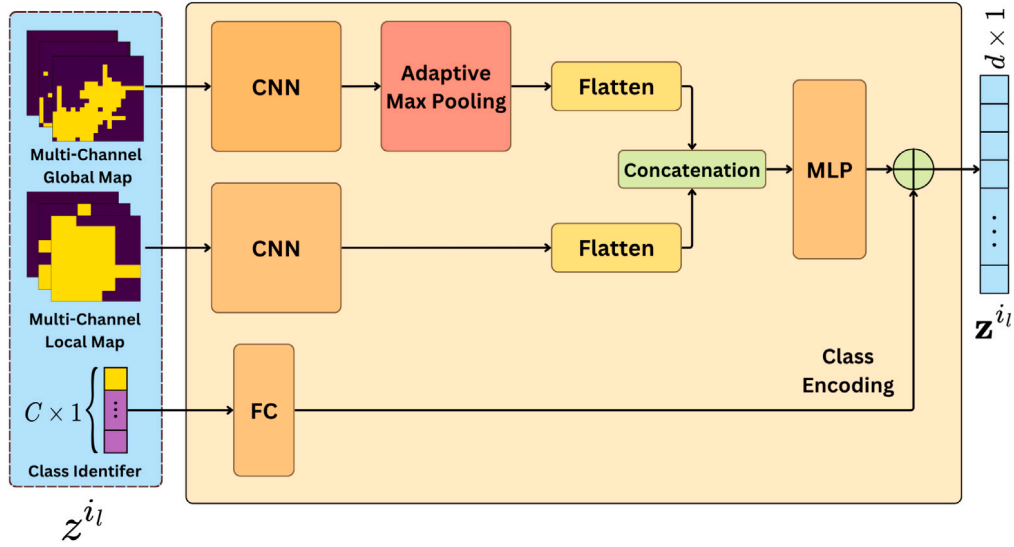

**Fig. 3.** The macro-observation embedder network, where agent $i_l$'s macro-observation $z^{i_l}$ is transformed into a macro-observation embedding $\mathbf{z}^{i_l} \in \mathbb{R}^d$.

Similar to the idea of agent indication [25], a learnable agent class encoding is used to enable unique behaviors for different classes of agents. This class encoding is obtained by processing a one-hot agent class identifier of size $C$ tied to each agent's MO $z^{i_l}$ through a fully connected layer (FC). In our case study with explorer and rescuer agents, $C$ is chosen as 2. However, more agent classes can be considered by simply increasing the size of the one-hot class identifier vector. The outputs of the MLP and FC layers are combined via element-wise addition, resulting in the macro-observation embedding $\mathbf{z}^{i_l}$.

### 5.3. Encoder

The encoder is made up of several encoding blocks each consisting of a self-attention mechanism, an MLP, and residual connections. It processes the sequence of MO embeddings $\{\mathbf{z}^{i_l}\}_{l=1}^{m}$ into a sequence of MO representations $\{\hat{\mathbf{z}}^{i_l}\}_{l=1}^{m}$, which carry information both about each agent's current view of the environment, as well as the high-level interrelationships among the agents. An additional MLP is also used during the training phase to approximate the value of each agent's macro-observation. Values associated with the active subset of agents, $(V_\phi^{i_1}(\hat{\mathbf{z}}^{i_1}), \ldots, V_\phi^{i_g}(\hat{\mathbf{z}}^{i_g}))$, are used to train the encoder and MO-embedder by minimizing the empirical Bellman error

$$L_{Encoder}(\phi) = \frac{1}{N}\sum_{i=1}^{N}\frac{1}{T^i}\sum_{\tau^i=0}^{T^i-1}\left[\mathcal{R}_{\tau^i}^{i} + \gamma V_{\bar{\phi}}^{i}(\hat{\mathbf{z}}_{\kappa^i(\tau^i+1)}^{i}) - V_{\phi}^{i}(\hat{\mathbf{z}}_{\kappa^i(\tau^i)}^{i})\right]^2 \quad (9)$$

to estimate the local macro-observation value function, where $T^i$ is the total number of MA updates of agent $i$, $\kappa^i$ is the mapping between local macro-step of agent $i$ and the global macro-step, $\phi$ represents MO-embedder and encoder parameters, and $\bar{\phi}$ represents the non-differentiable target network's parameters.

### 5.4. Macro-actions embedder

This module converts one-hot encoded representation of MAs $m^{i_{0:l-1}}, l = \{1, \ldots, m\}$ into MA embeddings $\mathbf{m}^{i_{0:l-1}}$ using an MLP. Similar to the MO-embedder, class encodings are combined with these embeddings to associate each MA with the agent class responsible for executing it.

### 5.5. Decoder

The decoder processes the joint MA embeddings $\mathbf{m}^{i_{0:l-1}}$, $l = \{1, \ldots, m\}$ through a series of decoding blocks, with $\mathbf{m}^{i_0}$ acting as an arbitrary token designating the start of decoding. Each decoding block is made up of a masked self-attention mechanism, a masked attention mechanism, and an MLP followed by residual connections. The masking ensures that each agent is only attending to itself and the agents preceding it, preserving the sequential updating scheme and the monotonic performance improvement guarantee during training [26]. The final decoder block outputs a sequence of joint MA representations $\{\hat{\mathbf{m}}^{i_h}\}_{h=0}^{l-1}$, which is then fed to an MLP to obtain the probability distribution of agent $i_l$'s MA, which is the high-level policy $\mu_\theta^{i_l}(m^{i_l}|\hat{\mathbf{z}}^{i_{1:m}}, \hat{\mathbf{m}}^{i_{0:l-1}})$, where $\theta$ represents the MA-embedder and decoder parameters. The decoder is trained by minimizing the following clipped PPO objective, which only uses the action probabilities and advantage estimates of the active subset of agents $i_{1:g(\tau)}$ at macro-step $\tau$:

$$L_{Decoder}(\theta) = -\frac{1}{T}\sum_{\tau=0}^{T-1}\frac{1}{g(\tau)}\sum_{l=1}^{g(\tau)}\min(r_\tau^{i_l}(\theta)\hat{A}_\tau^{i_l}, clip(r_\tau^{i_l}(\theta), 1 \pm \epsilon)\hat{A}_\tau^{i_l}), \quad (10)$$

$$r_\tau^{i_l}(\theta) = \frac{\mu_\theta^{i_l}(m_\tau^{i_l}|\hat{\mathbf{z}}_\tau^{i_{1:g}}, \hat{\mathbf{m}}_\tau^{i_{0:l-1}})}{\mu_{\theta_{old}}^{i_l}(m_\tau^{i_l}|\hat{\mathbf{z}}_\tau^{i_{1:g}}, \hat{\mathbf{m}}_\tau^{i_{0:l-1}})}, \quad (11)$$

where $\hat{A}_\tau^{i_l}$ is the estimate of agent $i_l$'s advantage function obtained using *generalized advantage estimation* (GAE) [33], based on the agent's individual reward sequence and $V_\phi^{i_l}(\hat{\mathbf{z}}_\tau^{i_l})$ as the local value function.

Actions are generated in an autoregressive manner during inference, which means that generating $m^{i_{l+1}}$ requires $m^{i_l}$ to be inserted back into the decoder. However, the output probability of all MAs $m^{i_{1:m}}$ can be computed in parallel during the training stage since $m^{i_{1:m-1}}$ have already been collected and saved in the replay buffer. The masked attention ensures that tokens for inactive agents do not influence the calculations for the active subset, and may simply be replaced with zero padding.

It is important to note that the monotonic improvement guarantee, based on sequentially maximizing the macro-advantage defined over the sum of all agents' expected returns (Eq. (6)), holds exactly only when agents share a fully cooperative (i.e., identical) reward function. In this work, the advantage function for each agent is estimated using GAE with respect to its own local rewards and value function estimate. As a result, the strict monotonic improvement guarantee does not always apply in the presence of heterogeneous reward functions, as local advantages may not perfectly align with the global objective. Nevertheless, in our setting, individual reward functions are specifically designed to support the team goal, so improvements in local advantages are expected to correspond to improvements in team performance in practice. This local estimation approach also enables efficient and scalable training in large heterogeneous teams, as it avoids the exponential complexity of maintaining a joint value function, while still conditioning each local value function on the joint macro-observations of the team.

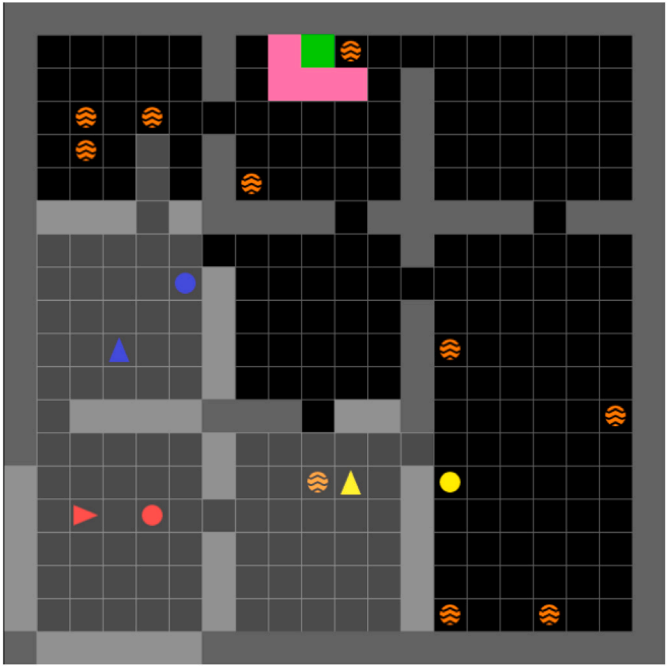

**Fig. 4.** A snapshot of a simulated episode in the Minigrid environment, where two explorer agents (yellow and blue triangles) and one rescuer agent (red triangle) navigate an unknown area in search of a target (green square). The pink squares are detectable as clues to the target's location. Agents' current navigation goals are marked by corresponding colored circles. (For interpretation of the references to color in this figure legend, the reader is referred to the web version of this article.)

## 6. Simulation setup

### 6.1. 2D simulation environment

CATMiP is evaluated in a search and rescue scenario with *explorer* and *rescuer* robot types, inside a customized 2D grid-world environment built using Minigrid [34]. Each episode features randomly generated environments of size $S \times S$, consisting of cluttered rooms with diverse shapes and a static target placed at an unknown location.

A snapshot of the simulation is shown in Fig. 4, where a rescuer robot (red triangle) and two explorer robots (blue and yellow triangles) navigate the environment to locate the target (green square). Each robot's global goal is marked by a corresponding colored circle. Cells are either free space (black) or occupied by walls or objects (orange circles), with explored areas visually highlighted. Free cells adjacent to the target are marked pink and serve as detectable clues for the robots. All robots have a $7 \times 7$ field of view in front of them that gets blocked by walls and obstacles, and they move at different speeds: explorer robots traverse a cell in one time step, while the rescuer robot takes two time steps for actions like moving forward or turning.

### 6.2. Reward structure

The main objective is for a rescuer robot to reach the target as quickly as possible. Each agent $i \in \mathcal{I}$ receives a time-dependent team reward $r^i_{success}(t) = 300(1 - 0.9\frac{t}{h})$ upon the mission's completion, incentivizing strategies that result in a quick rescue. Additionally, a similar time-dependent team reward $r^i_{locate}(t) = 100(1 - 0.9\frac{t}{h})$ is given to all agents at the time step the target is discovered. To conserve energy and reduce unnecessary movement, agents incur a small penalty of $r^i_{movement}(t) = -0.05$ whenever they move to a new cell.

Even though agents do not receive class-specific intermediate rewards in this scenario (such as exploration rewards for the explorer class), they still learn to adopt heterogeneous skills and behaviors that are suited to achieving the overall team objective.

### 6.3. Macro-action and macro-observation spaces

Global goal candidates, or macro-actions, are defined as cells within a square of side length $D$ centered on the agent, giving rise to a discrete macro-action space of size $D^2$. Once a macro-action is selected, it is mapped to its corresponding coordinate $(x, y)$ on the grid map, establishing the agent's global goal. Invalid actions, such as those targeting occupied or out-of-bounds cells, are handled using invalid action masking [35,36].

Each agent's macro-observation is composed of three elements: a global information map of size $S \times S \times 7$, a local information map of size $L \times L \times 6$, and a one-hot encoded agent class identifier. The channels in the local map provide information about cells in the agent's immediate vicinity. These channels indicate which cells have been explored, the cells' occupancy, the location of the target and clues around it, the agent's current navigation goal, the locations of other rescuer robots, and the location of other explorer robots. Similarly, the global map represents this information for the entire map, with an extra channel representing the agent's current location.

During centralized training or perfect communication conditions, these channels contain the latest information from other agents. However, during distributed execution when there is no communication link between explorer agent $i$ and rescuer agent $j$ for example, current location of agent $j$ would not show up in agent $i$'s rescuer agents' location channel and agent $i$'s location would not be visible in agent $j$'s explorer agents' location channel.

In Minigrid, the primitive action space includes four actions: moving forward, turning right, turning left, and stopping. An agent's primitive actions are selected to follow the shortest path to the agent's current navigation goal generated by the A* algorithm, avoiding obstacles and other robots.

## 7. Simulation results and analysis

In this section, we evaluate the performance of the proposed CAT-MiP framework in the described scenario across various environment sizes, team sizes and compositions, and communication constraints. We also analyze its sensitivity to errors in the occupancy map obtained via the C-SLAM module, and the effect of macro-action range and duration on performance. For this purpose, we also trained a synchronous variant of CATMiP (Synch-CATMiP) as a well as a model with a smaller macro-action space (CATMiP-SmallMA), and compare all models against various established planning-based multi-agent and single-agent baselines.

### 7.1. Training setup

We trained **CATMiP** with asynchronous centralized training on an NVIDIA GeForce RTX™ 3090 GPU. The training was conducted across 64 parallel environments with a map of side-length size $S = 20$ for a total of 62,500 episodes with an episode horizon of 200 steps. Each

**Table 2**

Hyperparameters used for all models unless otherwise noted.

| Parameter | CATMiP/Synch-CATMiP | CATMiP-SmallMA |
|---|---|---|
| Local Decision Range ($D$) | 7 | 5 |
| MA Space Size ($|\mathcal{A}|$) | 49 | 25 |
| Local Obs. Window ($L$) | 7 | 7 |
| Global Pool Range ($G$) | 4 | 4 |
| Embedding Size ($d$) | 192 | 192 |
| # Transformer Att Heads | 1 | 1 |
| # Transformer Blocks | 1 | 1 |
| Discount Factor ($\gamma$) | 1.0 | 1.0 |
| Learning Rate (initial) | $10^{-4}$ | $10^{-4}$ |
| LR Schedule | Linear decay | Linear decay |
| Max MA Duration | 10 | 10 |
| PPO Epochs | 10 | 10 |
| PPO Clipping Parameter ($\epsilon$) | 0.05 | 0.05 |
| Total Parameters | 1,280,786 | 1,271,546 |
| Optimizer | Adam | Adam |

episode included 3 agents: one rescuer agent, one explorer agent, and a third agent randomly assigned to either class. The asynchronous training process took approximately 285 h. We also trained two other variants of our model for comparison:

**Synch-CATMiP** was trained under the same conditions, but in a synchronous manner similar to MAT [26]. During training, macro-actions were updated for all agents simultaneously every 10 time steps. **CATMiP-SmallMA** was also trained under the same settings as asynchronous CATMiP, but with a smaller local decision-making range of $D = 5$, resulting in $D^2 = 25$ possible macro-actions. Due to fewer trajectories being collected and stored in the training buffer, training Synch-CATMiP over the same total number of episodes required 166 h. Training CATMiP-SmallMA took 303 h, likely due to the increased number of decision-making events resulting from the smaller macro-action space.

In CATMiP and CATMiP-SmallMA, a maximum duration of 10 time steps were set for each MA, after which agents would activate again following the delay mechanism explained in Section 5.1 with $(t_{min}, t_{max}) = (2, 5)$. In all three variants, a macro-action is also interrupted if the agent receives information of the target's location, or if the current navigation goal is found to be occupied. Notable hyperparameters used for training these models are presented in Table 2.

Fig. 5 shows the progression of the average mission success rate and the average agent reward throughout the training for all models, where an exponential moving average with a span of 200 applied for visualization. All three models show similar trends over 62,500 episodes, achieving high success rates early but continuing to improve as training progresses. However, both Synch-CATMiP and CATMiP-SmallMA consistently lag behind CATMiP in terms of success rate and average rewards during training, indicating that agents in these variants take longer to reach the target. This is expected: in Synch-CATMiP, agents must wait for the next MA update even if they finish their current one before the set duration, while in CATMiP-SmallMA, more frequent decision points lead to increased delays in agent activation.

It should be noted that training models in larger environments with more agents are possible but would require more memory and computation time. Although the time complexity of the search to obtain new MAs increases only linearly with the number of agents, the simulation engine also has to handle generation, storage, and alignment of more experience trajectories during the episodes, which include processing the sensor inputs and merging maps, forming MOs, performing A* search for navigation, etc. For example, while training CATMiP with 3 agents in an environment of size 20 × 20 took 16.4 s per episode, it takes approximately 50 seconds per episode to train the model on a 32 × 32 map with 6 agents.

### 7.2. Baseline methods

We compare our method with planning-based exploration methods, including the multi-agent method of artificial potential field (APF) [37] and three single-agent frontier-based methods, namely a utility-maximizing algorithm (Utility) [38], a search-based nearest frontier method (Nearest) [39], and a rapid-exploring-random-tree-based method (RRT) [40]. The single-agent methods are adapted to multi-agent settings by planning on the shared global map, and their implementation on the Minigrid environment is adapted from Yu et al. [20]. During communication dropouts, the shared global map contains only the latest information received from other agents. All agent classes were treated identically during exploration. However, once a rescuer agent detects the target's location, it immediately navigates along the shortest path to the target.

Direct comparison to synchronous MARL baselines (e.g., MAPPO) is omitted here due to fundamental differences in agent heterogeneity and reward structure, adaptability to training and evaluation with different team sizes, and the asynchronous and distributed execution process. Adapting existing MARL methods to our setting would require substantial modification and design choices, making comparisons ambiguous and potentially misleading. Moreover, the recent work by Wen et al. [26] has already established the superior performance of the sequential approach of MAT over other MARL baselines in various cooperative homogeneous and heterogeneous benchmarks. Since CATMiP builds upon MAT, our focus is on analyzing the benefits of our asynchronous, heterogeneous, and distributed decision-making framework.

### 7.3. Evaluation results and analysis

The trained CATMiP variants and the planning-based baselines were evaluated on three tasks with increasing complexity. Task 1 involved one rescuer and one explorer agent in a 15 × 15 grid. Task 2 increased the map size to 20 × 20, with an additional explorer agent in the team. Task 3 scaled up to a 32 × 32 environment with 6 agents consisting of 2 rescuers and 4 explorers. Experiments to evaluate the success rate of different conditions are performed over 100 randomized episodes with the same seed. Agents acted asynchronously during all evaluations. We found that the AMAT policy network can generate MAs with average inference times of 4ms, 5.2ms, and 9ms for 2, 3, and 6 input tokens (agent MOs), respectively. Thanks to its compact architecture and low computational requirements, CATMiP can be efficiently deployed on-board robots using embedded hardware, and scales well with increasing team size.

**Scalability and Performance Under Communication Loss:** Fig. 6 shows the success rate of the different methods against mission time in Task 1. This comparison is made for three cases with different communication constraints. In Fig. 6(a) the agents have consistent communication throughout the mission, whereas in (b) and (c) the value of $\sigma$ in Eq. (8) is set to 4 and 2 respectively, indicating increasing levels of communication loss. The learning-based models outperform the baselines in all three cases by showing higher success rates within the same time-frame. For example, as seen in Fig. 6(a), 97% of the experiments using the CATMiP variants were successful by the 100th step since the start of the mission, while the best performing baseline method, RRT, achieves 86% success by the same time. Since the models were trained for a more complex task, this shows the scalability of CATMiP to smaller environments and team sizes.

The same comparison between the models with different communication constraints is shown for Task 2 in Fig. 7. Once again, the learning-based models outperform the baselines in all three communication scenarios.

For Task 3, with a map size of 32 × 32 and 6 agents, the learning-based models still show top performance alongside the planning-based methods RRT and Nearest, as shown in Fig. 8. As the communication loss increases, our model's superiority becomes more prominent. Since the CATMiP variants were trained on a smaller map of size 20 × 20 and with a smaller team of agents, results on Task 3 show the scalability of our proposed framework to more complex tasks, with larger maps and team sizes. Moreover, the CATMiP model shows better adaptability and performance in asynchronous execution compared to the Synch-CATMiP and CATMiP-SmallMA variants in this large environment, especially as the communication loss increases and agents have to act mostly based on their own observations only.

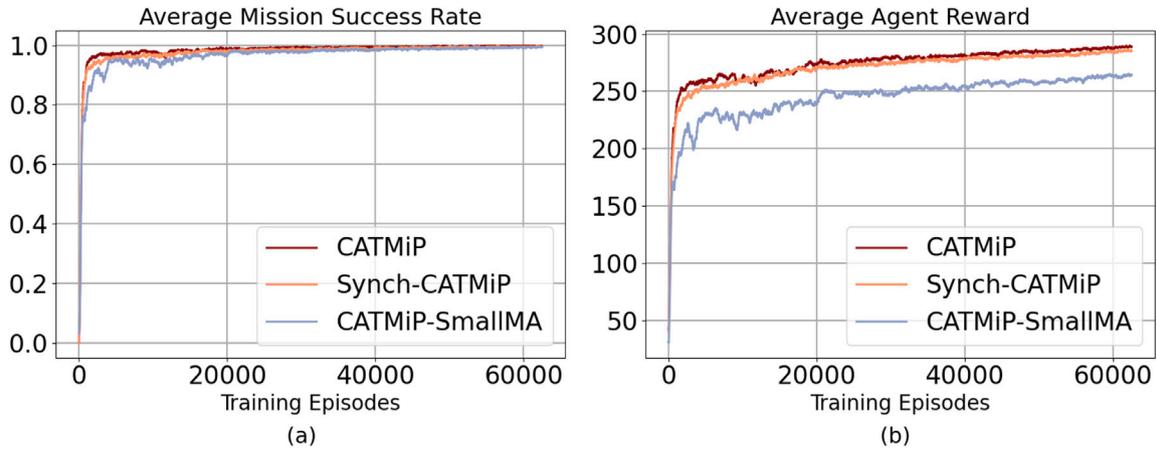

**Fig. 5.** (a) Progression of the mission success rate and (b) progression of average agent rewards against the number of episodes during training of the three models.

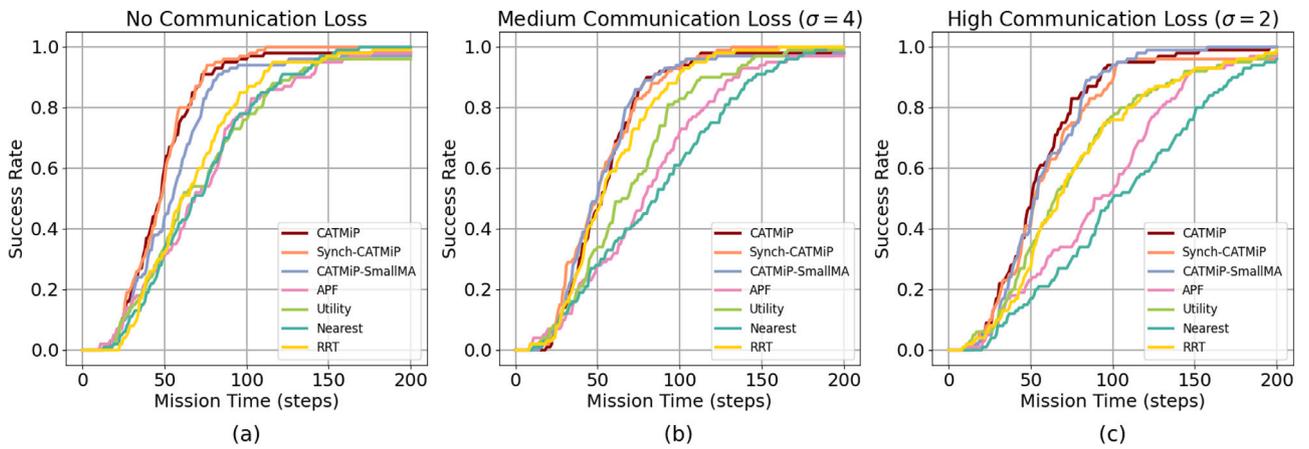

**Fig. 6.** Success rate of different models on Task 1 against mission time, with different communication constraints: (a) distributed execution with no communication loss, (b) distributed execution with moderate communication loss, and (c) distributed execution with heavy communication loss.

**Robustness to noisy C-SLAM:** To assess the impact of localization and mapping errors, we introduced noise to the occupancy grid by flipping the status of each explored cell with a 5% probability at every time step. Experiments were carried out on all three tasks under both ideal (consistent communication) and highly lossy ($\sigma = 2$) communication conditions. Fig. 9 shows that planning-based methods suffer a greater drop in performance compared to the CATMiP variants when given a noisy occupancy grid map. When both heavy communication loss and SLAM noise are present (Fig. 10), the gap between learning-based methods and planning-based methods widens further, as the compounded effects of map inconsistencies and delayed information disproportionately affect the planning baselines. These results highlight the ability of CATMiP to adapt and operate effectively even when provided with noisy and inconsistent map data.

**Effect of Macro-Action Range and Duration:** In all the evaluations above, agents followed the same maximum MA duration of 10 time steps that was also used during training. To further investigate the impact of macro-action design on performance, we evaluated all three model variants on Task 2 with no communication loss or map noise, but with different maximum MA durations: 5, 10, and 20 steps.

Fig. 11 shows the mission success rate as a function of mission time for each setting. For all models, a shorter maximum MA duration (5 steps) enables more frequent high-level decision making, leading to faster mission completion in most cases. This effect is most pronounced in CATMiP and Synch-CATMiP, where success rates rise more quickly with lower MA durations, since agents can react and re-plan more often when MAs are interrupted earlier. As the maximum MA duration increases to 20 steps, agents become committed to their current MA for longer, slowing their ability to adapt to new information, which results in a slower rise in success rate. The model with a smaller macro-action space, CATMiP-SmallMA, exhibits a different trend: although more frequent decisions (shorter MA durations) improve responsiveness, the reduced local range leads to increased delays in reaching distant goals, particularly when the MA duration is very short. As a result, performance at the strictest duration (5 steps) is slightly worse than at the default 10 steps, with the best trade-off between adaptability and progress observed at the intermediate MA duration.

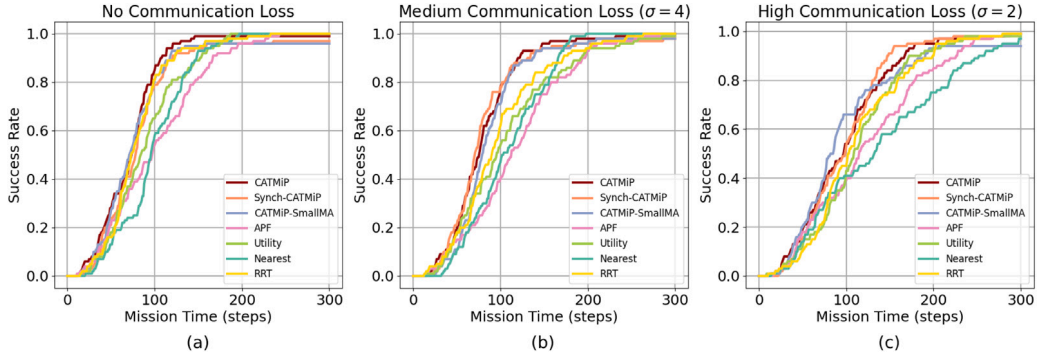

**Fig. 7.** Success rate of different models on Task 2 against mission time, with different communication constraints: (a) distributed execution with no communication loss, (b) distributed execution with moderate communication loss, and (c) distributed execution with heavy communication loss.

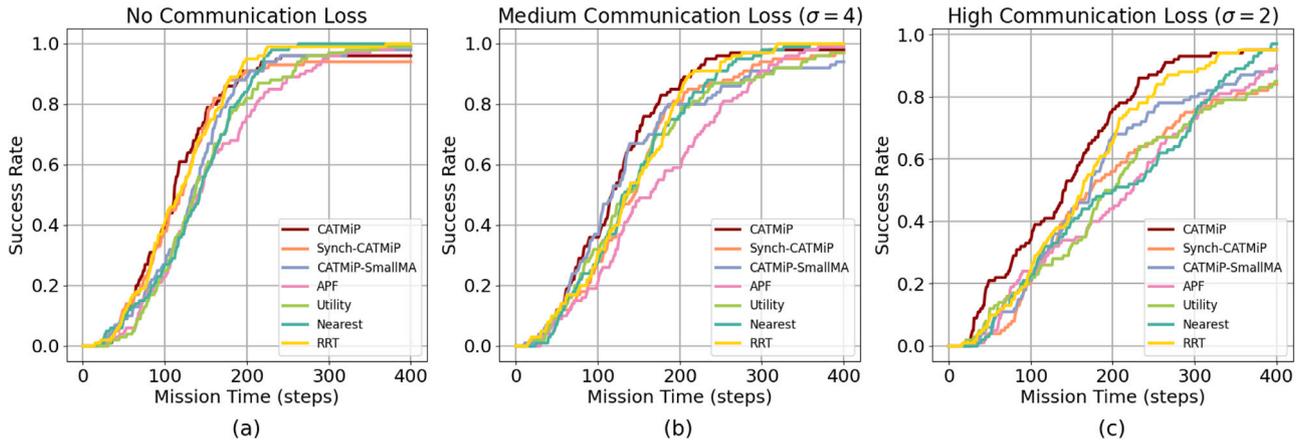

**Fig. 8.** Success rate of different models on Task 3 against mission time, with different communication constraints: (a) distributed execution with no communication loss, (b) distributed execution with moderate communication loss, and (c) distributed execution with heavy communication loss.

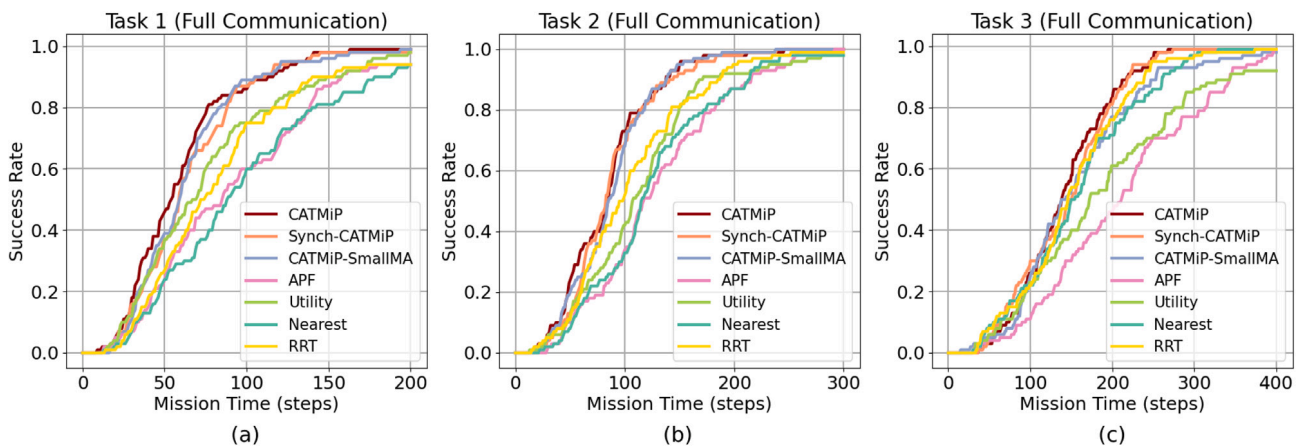

**Fig. 9.** Success rate of different models on all three tasks against mission time, with no communication loss and a 5% probability of occupancy value of explored cells being reported incorrectly.

These results highlight that both the spatial range of macro-actions and the frequency of high-level decision-making affect multi-agent coordination efficiency. While shorter MA durations generally improve adaptability and mission speed for larger action spaces, they can introduce inefficiency when combined with overly restricted macro-action ranges. Overall, CATMiP with its default range and moderate MA duration (10 steps) achieves the most robust performance across different settings.

## 8. Conclusions & Future work

This paper introduced CATMiP, a novel framework for coordinating heterogeneous multi-robot teams in environments with communication constraints. The proposed CMacDec-POMDP model provides the mathematical foundations of asynchronous and decentralized decision-making of heterogeneous agents by incorporating class-based distinctions across its components. Leveraging the transformer's ability to handle variable input sequence lengths, the Multi-Agent Transformer architecture was extended to develop scalable coordination strategies that can be executed in a distributed manner by any number of agents.

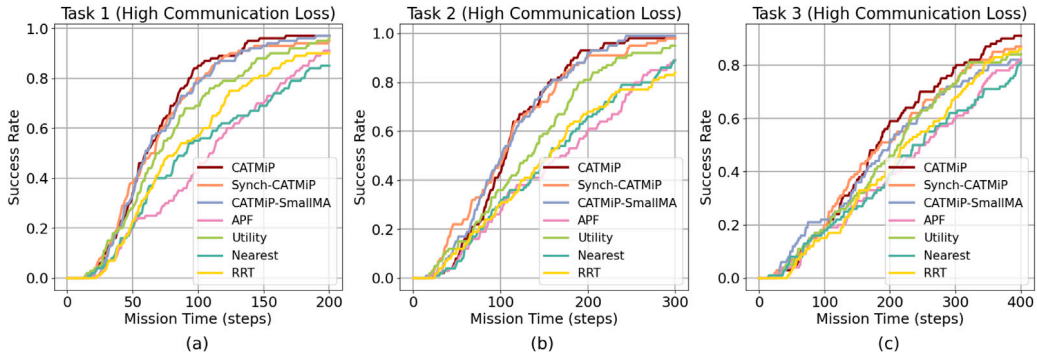

**Fig. 10.** Success rate of different models on all three tasks against mission time, with high communication loss ($\sigma = 2$) and a 5% probability of occupancy value of explored cells being reported incorrectly.

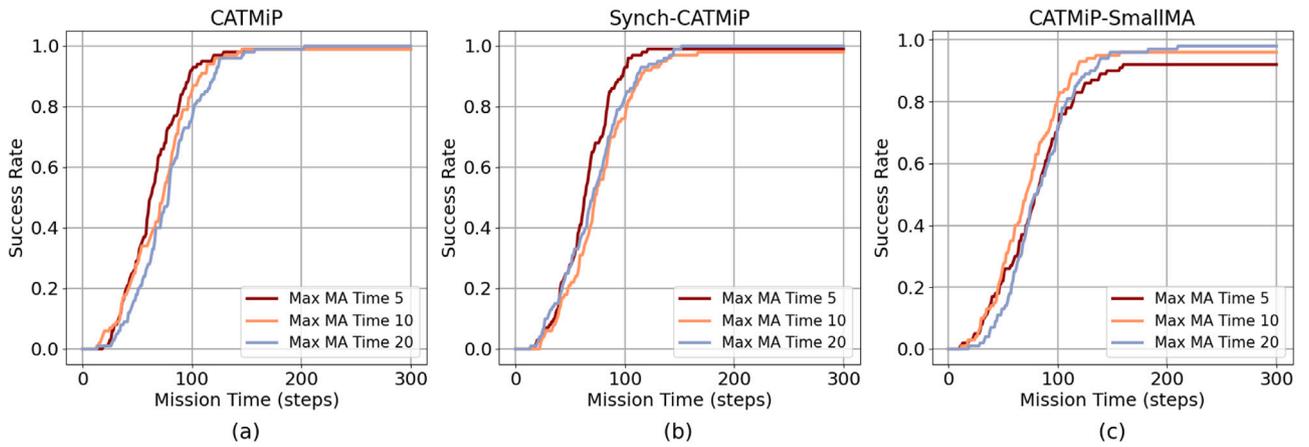

**Fig. 11.** Mission success rate on Task 2 as a function of mission time for three model variants, each evaluated with different maximum macro-action durations (5, 10, and 20 steps).

CATMiP demonstrated robustness and scalability across different team sizes, compositions, and environmental complexities in simulations of search and target acquisition tasks. The results highlighted the framework's adaptability to sporadic communication, asynchronous operations, and varied map conditions, achieving high mission success rates and competitive performance even under strict communication constraints. By addressing key challenges such as communication dropout, asynchronous operations, and agent heterogeneity, CATMiP shows significant potential for real-world applications where different types of mobile robots must cooperate under resource limitations and unpredictable conditions.

While our evaluation focused on 2D grid-world simulations, we have made several design choices to facilitate practical deployment on real robotic platforms. First, the inference time of the network is very fast (e.g., 9 ms with 6 agents), and the model is small enough to run efficiently on embedded hardware, making it suitable for real-time decision-making in multi-robot systems. Since each agent independently executes its own local copy of the model, the approach is inherently scalable to large teams, with no centralized computational bottleneck.

Nevertheless, several important deployment challenges remain. Communication delays and bandwidth limitations, which become more prominent in larger teams or more complex environments, need to be carefully addressed. This limitation is further exacerbated by the fact that agents must also share mapping and localization data for C-SLAM, increasing the overall communication load. While CATMiP is robust to sporadic communication (as demonstrated in our experiments), real-world networks may introduce additional latency, bandwidth constraints, and packet loss, which warrant further investigation in realistic or hardware-in-the-loop simulations.

Additionally, the gap between simulation and reality (sim2real) must be bridged for successful deployment. We have taken initial steps by experimenting with noisy SLAM outputs and introducing randomized actuator delays during training, improving the robustness of CATMiP to some real-world uncertainties. However, further work is required to account for unmodeled sources of error and variability encountered on physical robots, such as sensor noise, actuator drift, or hardware failures.

Future research will focus on evaluating the framework in more realistic 3D simulation environments and on physical robot platforms. We will also investigate integrating temporal memory into the AMAT network and adapting it to scenarios involving dynamic or adversarial targets. The framework's class-based architecture allows new agent types to be introduced and agents' behaviors to be customized through reward function design, highlighting its potential adaptability to more complex multi-robot applications. These efforts will further clarify

CATMiP's practical applicability and limitations, paving the way for robust deployment in real-world multi-robot systems.

**CRediT authorship contribution statement**

**Milad Farjadnasab:** Writing – review & editing, Writing – original draft, Visualization, Software, Resources, Methodology, Investigation, Data curation, Conceptualization. **Shahin Sirouspour:** Writing – review & editing, Supervision.

**Declaration of competing interest**

The authors declare that they have no known competing financial interests or personal relationships that could have appeared to influence the work reported in this paper.

**Data availability**

The codebase is shared as github repository in the abstract of the paper.

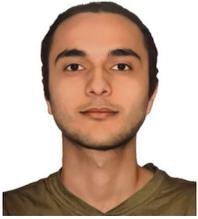

**Milad Farjadnasab** received the B.Sc. in Electrical Engineering in 2018, and M.Sc. in Control Engineering in 2020, both from Sharif University of Technology, Iran. He is currently a Ph.D. Candidate at the Department of Electrical and Computer Engineering at McMaster University, where his research is focused on deep multi-agent reinforcement learning for mobile robots.

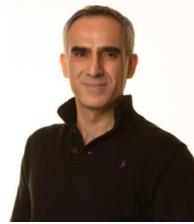

**Shahin Sirouspour** received the B.Sc. and M.Sc. degrees in electrical engineering from the Sharif University of Technology,Tehran, Iran, in 1995 and 1997, respectively, and the Ph.D. degree in electrical engineering from the University of British Columbia, Vancouver, BC, Canada, in 2003. He is currently a Professor with the Department of Electrical and Computer Engineering, McMaster University, Hamilton, ON, Canada. His research interests include autonomous systems, aerial robotics, teleoperation control, haptics, medical robotics, and optimization-based energy management and control in the smart grid environment.